\documentclass[letterpaper, 10 pt, conference]{ieeeconf}  

\IEEEoverridecommandlockouts                             

\usepackage{graphicx} 
\usepackage{amsmath}
\usepackage{bm}
\usepackage{xcolor}
\usepackage{lipsum}
\usepackage{amssymb}  

\usepackage[backend = biber, style=ieee]{biblatex}

\DeclareSourcemap{
  \maps{
    \map{
      \pertype{article}
      \step[fieldset=url, null]
      \step[fieldset=doi, null]
      \step[fieldset=issn, null]
      \step[fieldset=isbn, null]
      \step[fieldset=note, null]
      \step[fieldset=editor, null]
      \step[fieldset=urldate, null]
      \step[fieldset=file, null]
    }
  }
}
\DeclareSourcemap{
  \maps{
    \map{
      \pertype{inproceedings}
      \step[fieldset=url, null]
      \step[fieldset=doi, null]
      \step[fieldset=issn, null]
      \step[fieldset=isbn, null]
      \step[fieldset=note, null]
      \step[fieldset=editor, null]
      \step[fieldset=urldate, null]
      \step[fieldset=file, null]
    }
  }
}
\DeclareSourcemap{
  \maps{
    \map{
      \pertype{incollection}
      \step[fieldset=url, null]
      \step[fieldset=doi, null]
      \step[fieldset=issn, null]
      \step[fieldset=isbn, null]
      \step[fieldset=note, null]
      \step[fieldset=editor, null]
      \step[fieldset=urldate, null]
      \step[fieldset=file, null]
    }
  }
}

\title{\LARGE \bf
Optimization free control and ground force estimation with momentum observer for a multimodal legged aerial robot
}

\author{Kaushik Venkatesh Krishnamurthy$^{1}$, Chenghao Wang$^{1}$, Shreyansh Pitroda$^{1}$, \\ Eric Sihite$^{2}$, Alireza Ramezani$^{1*}$, Morteza Gharib$^{2}$ 
\thanks{$^{1}$The author is with Department of Electrical and Computer Engineering,
        Northeastern University, Boston, MA, USA  {\tt\small venkateshkrishnamu.k, wang.chengh, pitroda.s, @ northeastern.edu}}%
\thanks{$^{2}$The author is with the Department of Aerospace Engineering, California Institute of Technology, Pasadena, CA, USA {\tt\small esihite@caltech.edu}%
}
\thanks{$^{*}$Corresponding author {\tt\small a.ramezani@northeastern.edu}}}
\begin{document}

\maketitle
\thispagestyle{empty}
\pagestyle{empty}

\begin{abstract}
Legged-aerial multimodal robots can make the most of both legged and aerial systems. In this paper, we propose a control framework that bypasses heavy onboard computers by using an optimization-free Explicit Reference Governor that incorporates external thruster forces from an attitude controller. Ground reaction forces are maintained within friction cone constraints using costly optimization solvers, but the ERG framework filters applied velocity references that ensure no slippage at the foot end. We also propose a Conjugate momentum observer, that is widely used in Disturbance Observation to estimate ground reaction forces and compare its efficacy against a constrained model in estimating ground reaction forces in a reduced-order simulation of Husky. 
\end{abstract}

\section{Introduction}
Momentum observers have historically been extensively used in robot manipulators to detect collisions of unknown geometry and locations without sensors and were introduced in~\cite{de_luca_sensorless_2005, haddadin_robot_2017}. The method was attractive because of the ability to detect, isolate, and identify the forces using only proprioceptive sensors. Detection by using sensors and comparing torques against references that have a time-invariant threshold is usually very noisy because of the need to measure joint accelerations. Consequently, the theory of conjugate momentum observers has been implemented in legged robots for collision detection, isolation, and identification \cite{vorndamme_collision_2017}. The momentum observer is also preferred because of the ability to avoid the inversion of the mass inertia tensor and eliminating the need of an estimating the joint accelerations. 
\begin{figure}[t]
\includegraphics[width = \linewidth]{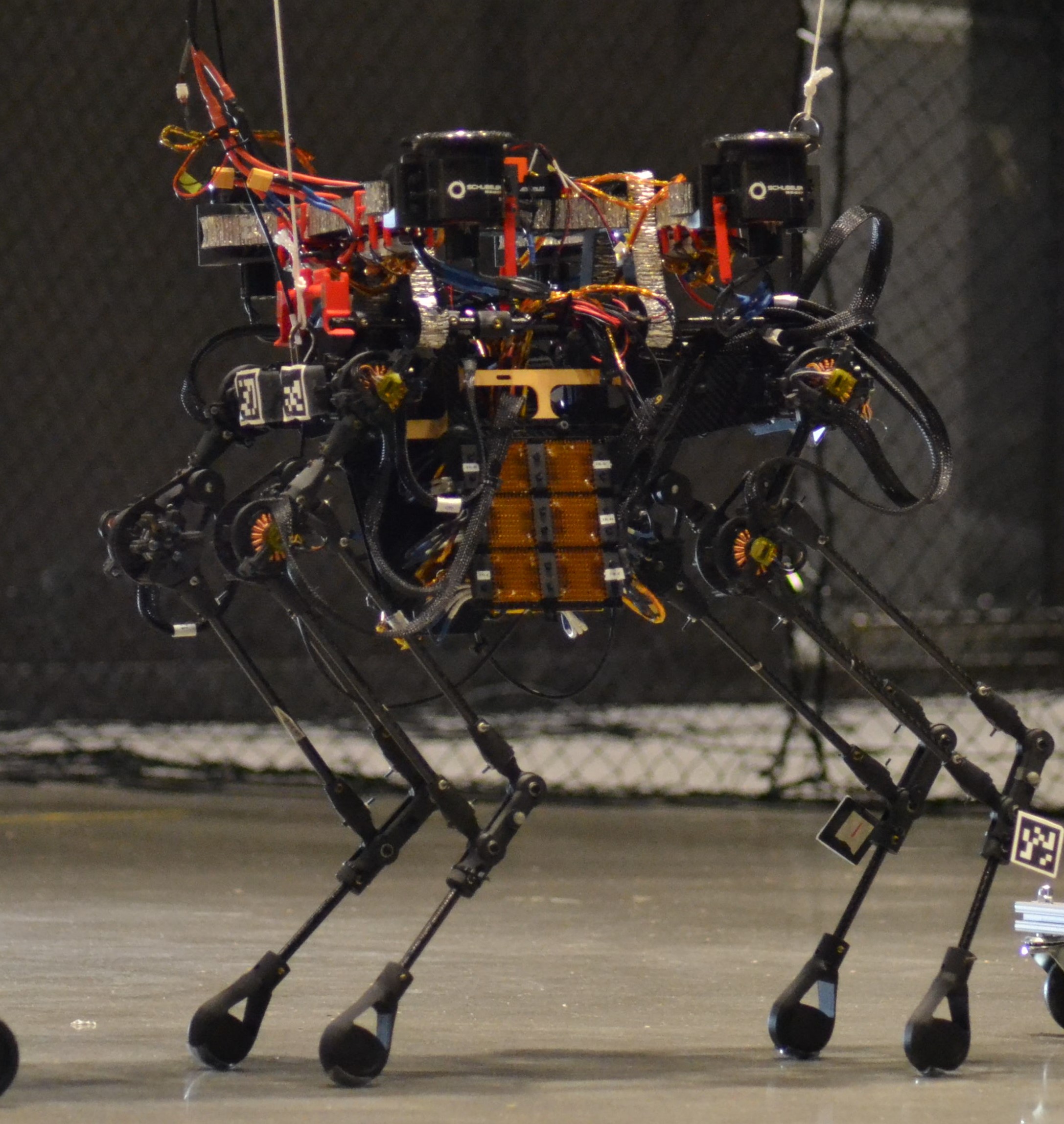}
\caption{Husky fitted with the Electric Ducted fans. Each leg is actuated with 3 BLDC motors with ELMO motor drivers}
\end{figure}

Multimodal robots \cite{salagame_quadrupedal_2023,dangol_control_2021,sihite_efficient_2022,sihite_multi-modal_2023} have the ability to exploit the advantages of the different modalities they possess. This allows the robots to not only exploit the best of each modality separately, but also repurpose to expand the boundaries of locomotion ~\cite{sihite_dynamic_2024, sihite_multi-modal_2023}. Multimodal legged-aerial systems have shown to have advantages over traditional legged robots and with the ability to perform thrust vectoring through posture manipulation. Ground force estimation then becomes critical for the performance of legged robots. 

With regards to legged-aerial systems, the momentum observer lends itself useful in estimating the ground reaction force wrenches. The wrench estimate can then be useful for different control strategies which define the inputs to the robot. Flacco et al. ~\cite{flacco_residual-based_2016} introduced an algorithm for contact estimation for humanoid robots using residual obtained from a momentum observer of a floating base model. The residual could also be potentially used for obtaining an estimate of the discrepancy between the actual robot dynamics and the the used model. Morlando and Ruggiero ~\cite{morlando_disturbance_2022} developed a `hybrid' observer by combining a momentum-based observer and an acceleration-based observer on a quadrupedal robot to estimate external disturbances.  Lim et al. \cite{lim_momentum_2021} use a momentum observer to estimate external disturbances and an LSTM to model the uncertainty with friction and modeling error Liu et al. \cite{liu_sensorless_2024} proposed a sensorless GRF observation method for a Heavy-Legged robot that can sustain upto 100kg of load using a sliding mode observer and non-linear disturbance observer that only uses the motor currents, speeds, and joint positions. Vorndamme et al. \cite{vorndamme_collision_2017} were able to detect, identify and isolate external collisions on an `Atlas' legged robot while compensating for ground contact forces using an ankle torque and foot contact sensors. The robot requires the presence of multiple sensors along the body for precise identification and isolation in the case of more than one collision force. 

Husky \cite{ramezani_generative_2021} is a multimodal legged aerial robot that has the ability use thruster forces from EDFs ( Electric Ducted Fans), attached to its torso, during legged locomotion. The EDFs can produce up to 2 kgf of thrust each. Husky stands 3 feet tall and is about 1.5 feet wide. The robot weighs 8 kg and is actuated with 4 3-DOF legs. The Hip Frontal joint allows to move the leg in the frontal plane and the hip-sagittal (HS) joint works in tandem with the knee (K) joint to maneuver the leg in the hip sagittal plane. All 12 joints on the robot are actuated by T-motor Antigravity 4006 brushless motors, with the motor output transmitted through a Harmonic drive. The Harmonic drives are chosen for their precise transmission, low backlash, and back-drivability. The motor and gearbox in the joints were embedded in a custom 3D-printed housing during the printing process making the robot's legs significantly lightweight.

Through Husky Carbon, we want to push the boundaries of standard legged locomotion. Some previously done work is shown in \cite{salagame_quadrupedal_2023, krishnamurthy_narrow-path_2024}. Legged robots have the ability to manipulate ground reaction forces through posture manipulation and foot placement, limiting their operational capabilities. The onboard computing includes a RealTime machine for low-level motor control of the 12 motors which is facilitated through the Matlab Simulink model. For the purposes of high-level path planning and thrust control, the robot is equipped with a Pixhawk flight controller and an Nvidia Jetson Nano. 

Many of today's approaches to walking controllers through reduced order models depend on optimization solvers. Optimization solvers used in optimal controllers usually require powerful computers that can usually increase the payload of the robot which is not desired considering the conflicting requirements of legged-aerial robots.

This paper implements a modified explicit reference governor (ERG), walking controller for walking while also incorporating thruster forces applied from an attitude controller. Reference governors introduced in \cite{bemporad_reference_1998,gilbert_nonlinear_1994,garone_explicit_2016}, and presented in \cite{sihite_optimization-free_2021, sihite_efficient_2022, liang_rough-terrain_2021,dangol_towards_2020,dangol_hzd-based_2021} have been implemented successfully where the controller states are manipulated to satisfy Ground Reaction Force (GRF) constraints. in this work, the reference governor modulates the applied velocity reference to the body to avoid slippage of the feet. It assumes the robot to be a triangular inverted pendulum with a point mass, which is a suitable model to consider during a 2-point contact gait. We also propose a framework for, and show the effectiveness of,  a momentum observer used on HROM (Husky Reduced order Model) to estimate ground contact forces. 

To compensate for roll and pitch errors, thruster forces in the form of an external wrench.  Finally, a momentum-based estimator is used to estimate the ground reaction forces given the thruster inputs to individual thrusters.
\begin{figure}[t]
\includegraphics[width = \linewidth]{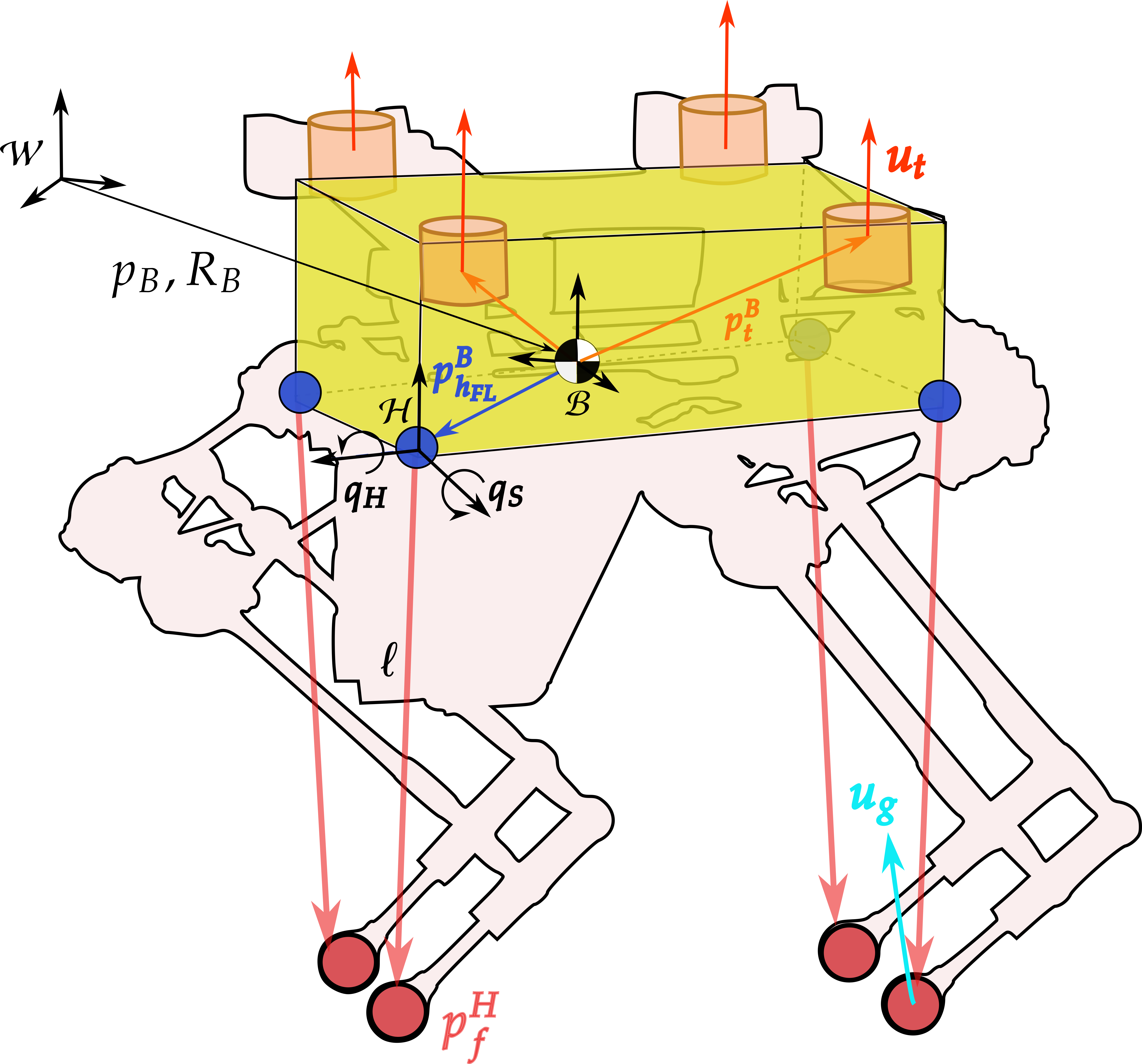}
\caption{ Depiction of the robot and HROM parameters. $\bm p_{h,i}, \bm p_{t,i}$ and $l_i$ are the locations of the hip joint, thruster position and the leg length of the $i^{th} $ leg/location}
\label{fig : Silhouette}
\end{figure}

\section{Modelling}
The equations of motion of the HROM can be derived using the energy-based Euler-Lagrange dynamics formulation. As shown in Fig.~\ref{fig : Silhouette}, the positions of the leg ends are defined as functions of the spherical joint primitives, namely $\bm q_B$ and $\bm q_H$, along with the length of the leg $l$. The pose of the body can be defined using $\bm{p}_B \in \mathbb{R}^3$, and Z-Y-X Euler angles $\bm{\Phi}_B$. The rotation matrix can also then be defined from the Euler matrix as $\bm R_B$. The generalized coordinates of the robot body can then be defined as follows:
\begin{equation}
\bm{q} = [\bm{p}_B^\top ,\bm{\Phi}_B^\top ]^\top,
\label{eq: body states}
\end{equation}
and the leg states of the robot can be defined as,
\begin{equation}
\begin{aligned}
\bm{q}_L &= [\dots, q_{H_i},q_{S_i},\ell_i, \dots]^\top, \\
\forall i &\in \mathcal{F},
\label{eq: leg states}
\end{aligned}
\end{equation}
where, $\mathcal{F} = [FR, FL, BR , BL]$ is the set containing all legs (front/back and left/right combinations). The position of the foot can then be determined using the forward kinematics equations shown:
\begin{equation}
\begin{aligned}
    \bm{p}_{F i} &= \bm{p}_{B} + \bm R_{B} \bm{p}_{h i}^{B} + \bm R_{B} \bm R_{y}\left(\phi_{i}\right) \bm R_{x}\left(\gamma_{i}\right)
    \begin{bmatrix} 
    0, & 0, & -\ell_{i}
\end{bmatrix}^\top
\label{eq:foot_pos}
\end{aligned}
\end{equation}
Let $\omega_B$ be the body angular velocity vector in the body frame and $g$ denote the gravitational acceleration vector. The legs of HROM are massless, so we can ignore all leg states and directly calculate the total kinetic energy $\mathcal{K}= \frac{1}{2}m \dot p_B^{\top} \dot p_B + \frac{1}{2}\omega_B^{\top} I_B \bm \omega_B$ (where $m$ and $I_B$ denote total body mass and mass moment of inertia tensor). The total potential energy of HROM is given by $\mathcal{V}=-m \bm p_B^{\top} g$. Then, the Lagrangian $\mathcal{L}$ of the system can be calculated as $\mathcal{L} = \mathcal{K} - \mathcal{V}$. Hence, the dynamical equations of motion are derived using the Euler-Lagrangian formalism. 

The body orientation is defined using Hamilton's principle of virtual work and the modified Lagrangian for rotation dynamics in SO(3) to avoid using Euler rotations which can become singular during the simulation. The equations of motion for HROM are given by
\begin{equation}
\begin{gathered}
    \textstyle \frac{d}{d t} \left( \frac{ \partial \mathcal{L}}{\partial \dot p_B } \right ) - \frac{\partial \mathcal{L}}{\partial p_B} = f_{gen}, \qquad
    \dot{\bm R}_B = \bm R_B\,[\omega_B]_\times \\
    \textstyle \frac{d}{dt}\left( \frac{\partial \mathcal{L}}{\partial \omega_B}  \right) + 
    \omega_B \times \frac{\partial \mathcal{L}}{\partial \omega_B} + 
    \sum_{j=1}^{3} \bm r_{B_j} \times \frac{\partial \mathcal{L}}{\partial \bm r_{B_j}} = \tau_{gen},
\end{gathered}
\label{eq:euler-lagrangian}
\end{equation}
where $f_{gen}$ and $\tau_{gen}$ are the the generalized forces and moments (from GRF and thrusters), $[\,\cdot\,]_\times$ is the skew operator, and $\bm R_B^\top = [\bm r_{B_1}, \bm r_{B_2}, \bm r_{B_3}]$ (i.e., $\bm r_{B_j}$ are the columns of $\bm R_b$).  The dynamic system accelerations can then be solved to obtain the following standard form:
\begin{equation}
\begin{gathered}
    \bm M(\bm{q})\dot{\bm{v}} + \bm h = \Sigma_{i \in \mathcal{F}} \left[ \bm B_{gi}\bm{u}_{gi} \right] + \bm{u}_t \\
    \bm B_{gi} = \frac{\partial{\dot{\bm{p}}_{f,i}}}{\partial{\bm{v}}},
\end{gathered}
\label{eq:manipulator eq}
\end{equation}

\noindent where $\bm M(q) $ is the mass-inertia matrix, $\bm h$ contains the Coriolis, and gravitational vectors, $\bm v = [\dot{\bm p_b}^\top, \bm{\omega_b}^\top]^\top$, and $\bm B_{gi} \bm u_{gi}$ represent the generalized force due to the GRF (Ground Reaction Forces) $\bm u_{gi}$ acting on the foot $i$. The term $\bm{u}_t \in \mathbb{R}^6$ $\bm u_t$ represents the actions exerted by the four thrusters, which are formed by condensing them into a wrench. The individual thruster forces are modeled as forces that can act only upwards in the body frame. 

The legs are driven by setting the joint variable accelerations to track desired joint states. The joint inputs are defined as follows
\begin{equation}
    \ddot{\bm q}_L = \bm u_L,
    \label{joint-inputs}
\end{equation}
\noindent where $\bm{u}_L$ forms the control input to the system in the form of the leg joint state accelerations. 
The full system of equations can then be derived from equation \ref{eq:manipulator eq} and equation \ref{joint-inputs} as follows:
\begin{equation}
\begin{gathered}
    \dot{\bm{x}} = \bm{f}(\bm{x},\bm{u}), \\
        \bm{x} = [\bm q_d^\top ,\bm{v}_d^\top ]^\top\\
        \bm{u} = [\bm{u}_t^\top, \bm{u}_L^\top]^\top\\
    \end{gathered}
\label{eq:eom}
\end{equation}
where $\bm q_d = [\bm q^\top, \bm q_L^\top]^\top, \bm{v}_d = \left[\bm{v}^\top, \bm{\dot{q}}_L^\top\right]^\top$, and $\bm{x}$ is obtained by combining both the dynamic and massless leg states and their derivatives to form the full system states. Finally, $\bm{u}$ is a vector of all inputs, which include the thrust wrench and the leg inputs. 
The GRF is modeled using a compliant ground model and Stribeck friction model, defined as follows:
\begin{equation}
\Sigma_{GRF}:\left\{\begin{aligned}
    \bm u_{gi} & = \begin{cases} \, 0 &  \mbox{if } z_i > 0  \\
    \, [u_{gi,x},\, u_{gi,y},\, u_{gi,z}]^\top & \mbox{else} \end{cases} \\
    u_{gi,z} &= -k_{gz} z_i - k_{dz} \dot{z}_i \\
    u_{gi,x} &= - s_{i,x} u_{gi,z} \, \mathrm{sgn}(\dot{x}_i) - \mu_v \dot{x}_i \\
    s_{i,x} &= \left(\mu_c - (\mu_c - \mu_s) \mathrm{exp} \left(-|\dot{x}_i|^2/v_s^2  \right) \right),
\end{aligned}\right.
\label{eq:ground-model}
\end{equation}
where $x_i$ and $z_i$ represent the $x$ and $z$ positions of foot $i$, respectively. $k_{gz}$ and $k_{dz}$ are the spring and damping coefficients of the compliant surface model, respectively. $u_{gi,x}$ and $u_{gi,y}$ denote the ground friction forces in the respective directions. $\mu_c$, $\mu_s$, and $\mu_v$ stand for the Coulomb, static, and viscous friction coefficients, respectively, and $v_s > 0$ represents the Stribeck velocity.
\section{Control}
\begin{figure}[t]
    \centering
    \includegraphics[width =
    \linewidth]{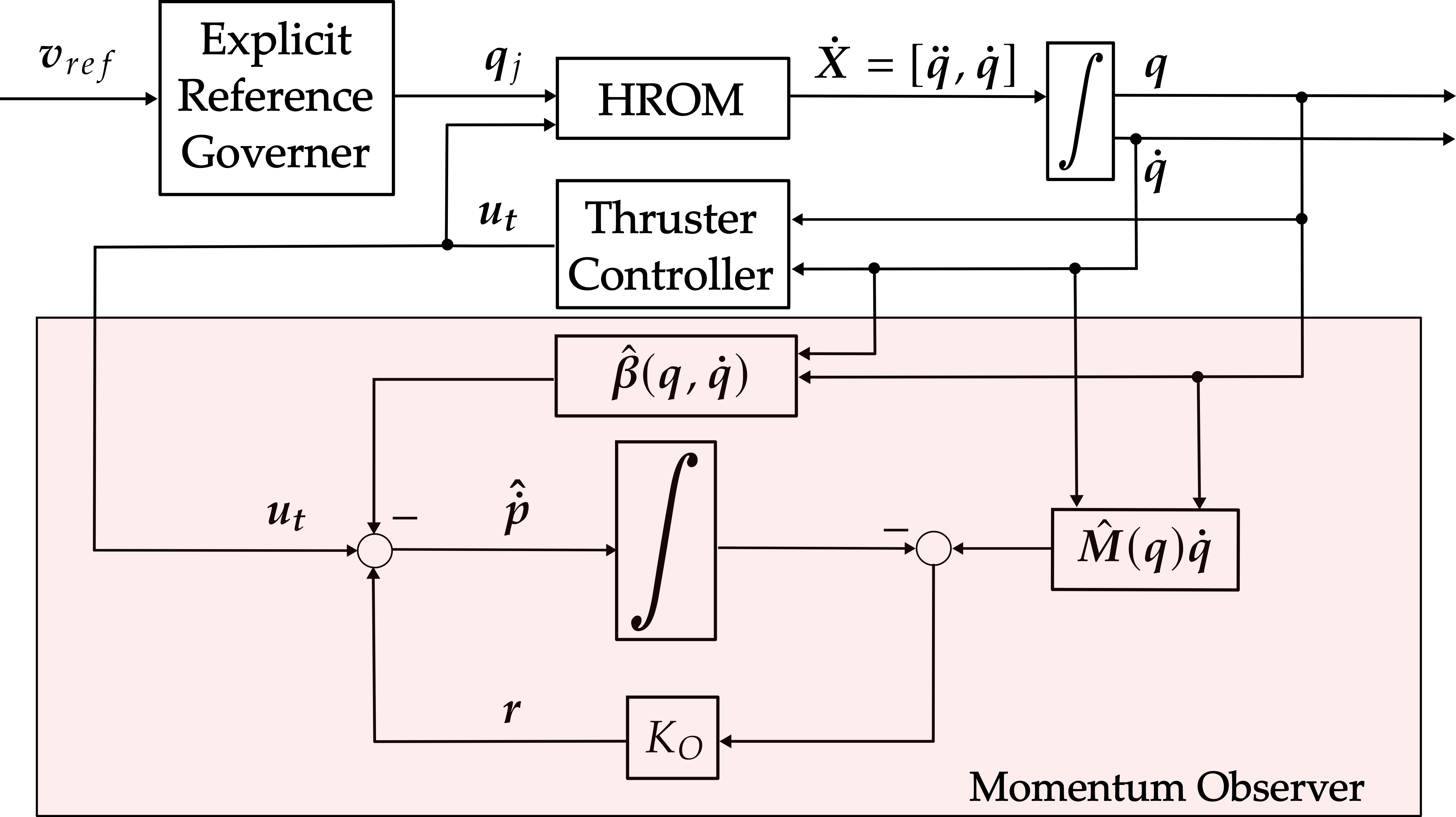}
    \caption{Flowchart of Control and estimation framework. The momentum observer has access to the control input from thruster attitude controller and the state of the simulation}
    \label{fig : flowchart}
\end{figure}

Fig. ~\ref{fig : flowchart} shows a flowchart of the control framework. The ERG filters the applied reference and outputs the joint trajectories for the filtered velocity. The HROM model shown in \ref{eq:eom} also takes the thruster commands from the PID controller. The observer has access to the robot states, velocities after integrating the state derivatives, thruster inputs, and uses the model parameters to calculate $ \hat{\bm{M}}$ and $\bm{\beta}$. The momentum $\bm p$ is initialized as zero at the start of the simulation. 
\subsection{Explicit Reference Governor}
\begin{figure}
    \centering
    \includegraphics[width=\linewidth]{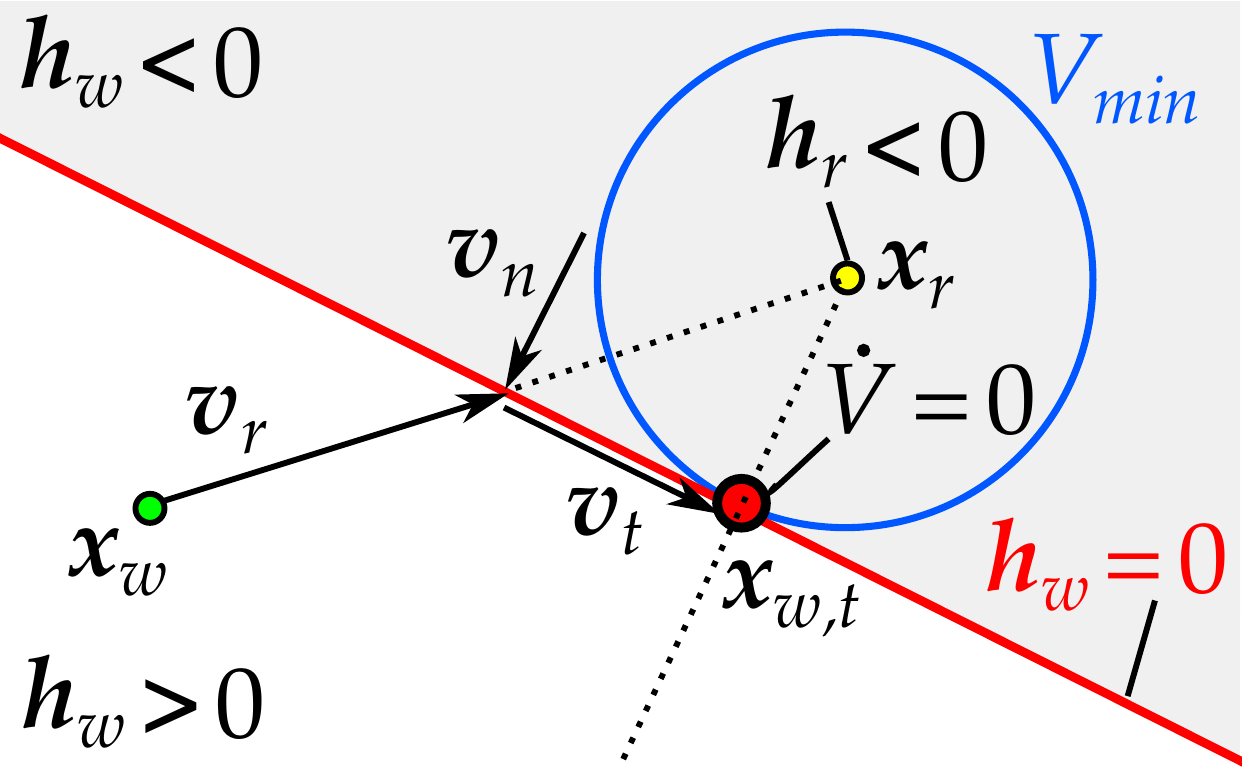}
    \caption{Explicit Reference Governor (ERG) manipulates the applied reference states \(\bm x_w\) to be as close as possible to the desired reference \(\bm x_r\) without violating the constraint equation \(\bm h_w\geq 0\). For the full algorithm refer to \cite{sihite_optimization-free_2021} from Sihite et al.}  
    \label{fig: erg_algorithm}
\end{figure}

\begin{figure}
    \centering
    \includegraphics[width=\linewidth]{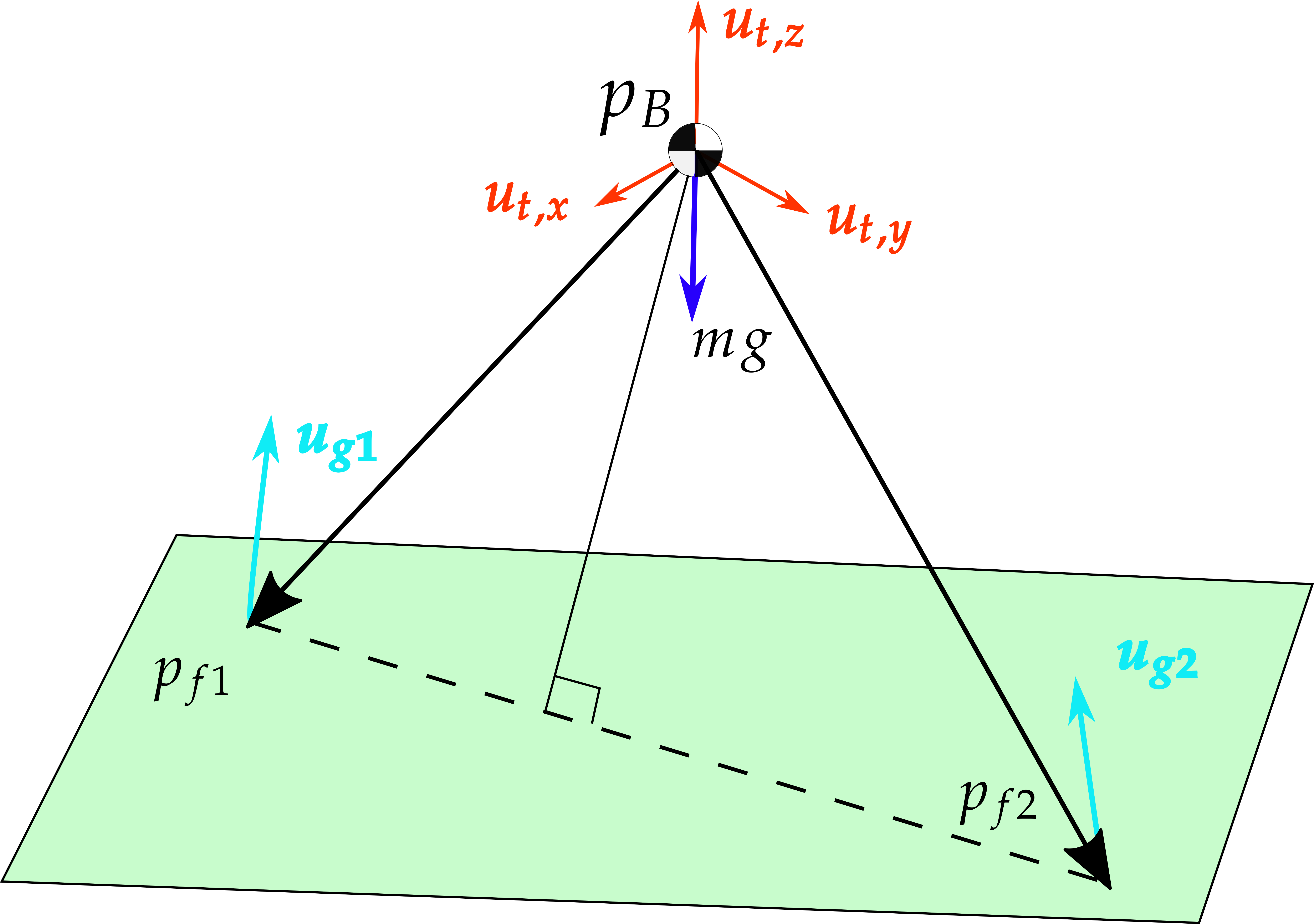}
    \caption{Model used in the ERG controller to estimate the ground reaction forces. The ERG then manipulates the reference to make sure the new applied reference is within the constraint admissible set}
    \label{fig:ERG_model}
\end{figure}

The Explicit Reference Governor is an add on a stabilized closed loop system. The Governor manipulates the applied reference to the controller to enforce the desired constraints while being as close to the desired reference trajectory. The ERG works using a provable Lyapunov stability properties and can tackle the problem in the state space in a much faster way using a relaxed inverted triangular inverted pendulum as shown in Fig.~\ref{fig:ERG_model}. A visual representation of this algorithm is shown in Fig.~\ref{fig: erg_algorithm}

The system of equations used to formulate the ground reaction forces 
\begin{equation}
    m_B\ddot{\bm p}_B = m_B \bm g + \bm u_{g1} + \bm u_{g2} + \bm u_t, 
\end{equation}
where $\bm u_{g1}$ and $\bm u_{g2}$ are forces in the front and rear legs during a 2 point contact gait. It can be assumed that the lateral ground forces are distributed evenly and the moment about the axis perpendicular to the support line is restricted. 
Using these equations, we can use yield a general system of equations where \(A [u^{\top}_{g1}, u^{\top}_{g2}]^{\top} = \bm b\). The constraint for the ERG is then calculated as follows, 
\begin{equation}
\bm h_r = \underbrace{\begin{bmatrix}
    \textrm{-sgn}(u_{gi,x}) & 0 & \mu_s\\
      0 & \textrm{-sgn}(u_{gi,x}) & \mu_s \\
       0 & 0 & 1
    \end{bmatrix} \bm u_{g,i}}_{\bm J_r \bm x_r} + \underbrace{\begin{bmatrix}
        0 \\ 0 \\ - u^{min}_{gi,z}
    \end{bmatrix}}_{\bm d_r} \geq 0
\end{equation}
Now using the formulation presented in Sihite et al. \cite{sihite_optimization-free_2021} the reference trajectory is updated based on the applied reference and the constraint. The ERG framework is utilized to enforce the friction pyramid constraint. 
\subsection{Attitude controller}

The attitude controller considers a simplified PID controller which reacts to errors on the roll pitch and yaw errors. 

\begin{equation}
    \bm u_t = \bm K_p(\bm{\Phi_{B,ref}} - \bm{\Phi_B}) + \bm K_d (\dot{\bm\Phi}_B),
\end{equation}
where $K_p$ and $K_d$ are PD gains for this controller. The output of the thruster forces are saturated to maximum allowable limit of the EDFs
\section{Estimation}



\subsection{Conjugate Momentum Observer}
 The generalized momentum of a system can be defined as \( \bm{\hat p} = \hat{\bm{M}} \hat{\dot{\bm q}}\) , with the \( \hat{\bm{(.)}} \) indicating an estimated value. Let us assume that we have access to $\bm u_t$ from the flight controller PWM inputs. For this, we also assume that the thrusters have been accurately characterized and there is a one to one mapping to the PWM input and the thrust obtained considering a stable nominal operating voltage from a power supply. The momentum observer dynamics of the system is then defined as, 
\begin{align*}
    \centering
    \hat{\dot{ \bm p}} &= \hat{\bm{M}} \hat{\ddot{\bm q}} + \hat{\dot{ \bm M}}\hat{\dot{\bm q}} \\
    &= \hat{\bm u}_g + \bm{B}_t\bm u_t -\hat{\bm h} + \hat{\dot{ \bm M}}\hat{\dot{\bm q}} \\
    &= \hat{\bm r } + \bm{B}_t\bm u_t - \hat{\bm{\beta}} \\
    \dot{ \bm r} &= \bm K_O \left( \dot{\bm{p}}(t) - \hat{\dot{\bm p}}(t) \right)\\
\end{align*}
Integrating the above equation gives us,
\begin{align*}
    \centering
      \bm r &= \bm K_O  \left( \bm p(t) - \int_0^t (\bm r - \hat{\bm {\beta}} + \bm B_t \bm u_t) dt \bm \right), \\
\end{align*}
where $\bm r$ here is the residual vector, $\bm K_O$ is a diagonal matrix of observer gains, $\hat{\bm{\beta}} = \hat{\bm h} - \hat{\dot{ \bm M}}(\bm q)\hat{\dot{\bm{ q}}} $. If we consider that, $\hat{\dot{\bm M}}(\bm q) = \dot{ \bm M}(\bm q)$. Numerically ${\dot{ \bm M}}(\bm q) $ can be calculated as $\frac{\bm M_k - \bm M_{k-1}}{T_s}$ and $\bm h$ is obtained from the derived Lagrange formulation of the dynamics using the Matlab symbolic Toolbox.
From Haddadin \cite{haddadin_robot_2017} then,
\begin{equation}
\bm K \rightarrow \infty \implies \bm r \approx \bm u_g 
\end{equation}
 Individual ground reaction forces from each leg could then be found by inverting the mapping matrix that maps the forces to the generalized coordinates. A flowchart of the simulation and the control framework is shown in Fig.~\ref{fig : flowchart}
\subsection{Constrained model estimation}
This estimated ground reaction force wrench is compared to the one obtained from a constrained ground model.  The constraint is written as, 
\begin{equation}
    \bm J \ddot{\bm q} + \dot{\bm J} \dot{\bm q} = \bm 0,
\end{equation}
where $\bm J$ is the matrix of stacked foot contact Jacobians. The constraint is formulated such that the feet are fixed to the stance foot locations with zero acceleration. The constrained ground reaction forces wrench is then found as follows, 
\begin{equation*}
    \bm {\lambda} = \left( \bm J \bm M^{-1} \bm J^{\top} \right)^{\dagger}\left( \bm J \bm M^{-1} (\bm u_t - \bm h) - \bm{\dot{J}}\bm{\dot{q}} \right),    
\end{equation*}
where $(.)^{\dagger}$ is the Moore-Penrose pseudo-inverse. During a two point contact gait, it is observed that the Delassus decoupling matrix  \(\bm J \bm M^{-1} \bm J^{\top}\) is not full rank and this makes the estimation of the ground forces inherently inaccurate. 
\begin{figure}[t]
\includegraphics[width = \linewidth]{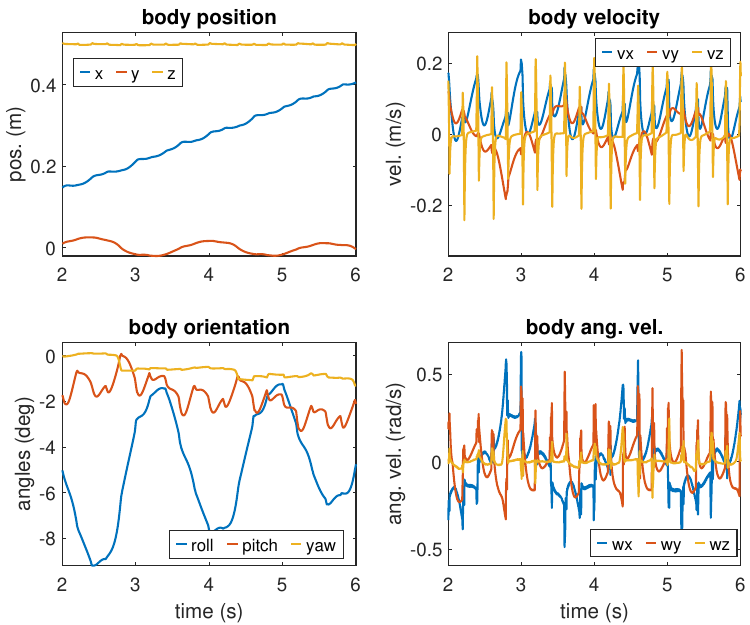}
\caption{Plots of robot body states with an applied reference for linear velocity of $v_{x,d} = 0.2 m/s$ and $\bm \Phi_B = \bm{0}_3$}
\label{fig : body_states}
\end{figure}
\section{Results}
\begin{figure}[t]
\includegraphics[width = \linewidth]{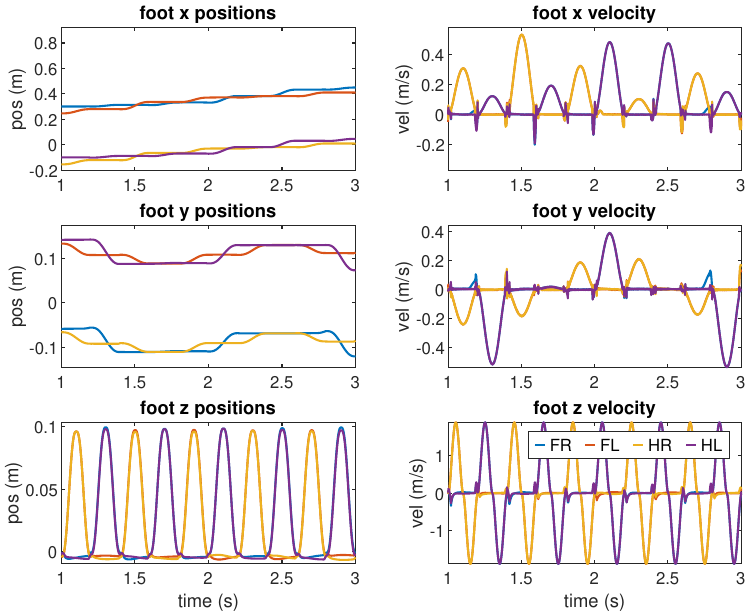}
\caption{Plots of foot end positions during simulations. }
\label{fig : foot positions}
\end{figure}
The simulation was performed on Matlab environment using a computer with an Intel core i7 processor and utilized the HROM framework. A fourth order Runge Kutta integrator was used to march the ODE forward. Basic heuristic were then applied to determine nominal gaits for a straight path, considering a specified forward velocity and step time, for up to 10 seconds. The walking pattern adopted a two point contact, where diagonally opposite leg pairs are synchronized while the remaining are operated out of phase. Each simulation of the time step includes the ERG computation, ODE integration, estimation with 4th order Runge-Kutta algorithm which was computed at a rate of approximately 2 kHz. The onboard computer SpeedGoat Realtime machine uses a C/C++ which are significantly faster than Matlab. 
\begin{figure}[t]
\includegraphics[width = \linewidth]{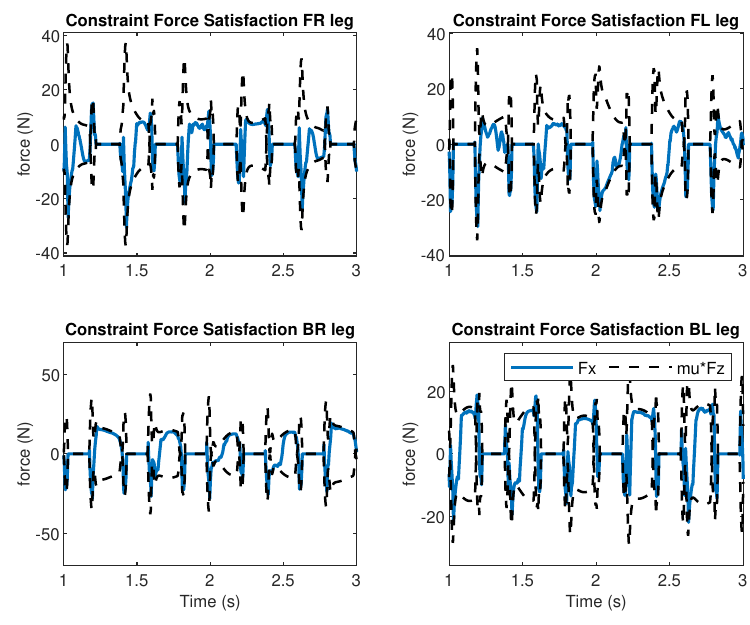}
\caption{Plots of constraint satisfaction with mu = 0.25. }
\label{fig : Friction cone constraint satisfication}
\end{figure}
\begin{figure}[t]
\includegraphics[width = \linewidth]{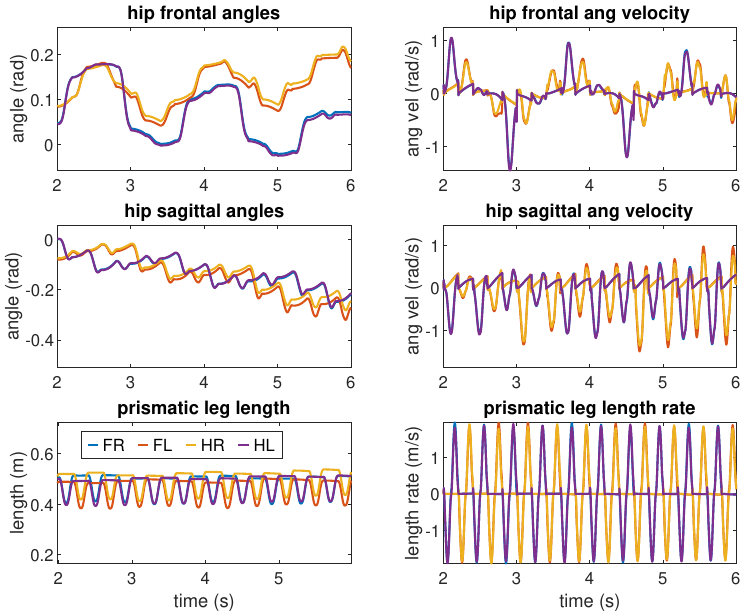}
\caption{Plots of leg angle and lengths during simulations }
\label{fig: leg angles}
\end{figure}
%
\begin{figure}[h]
\includegraphics[width = \linewidth]{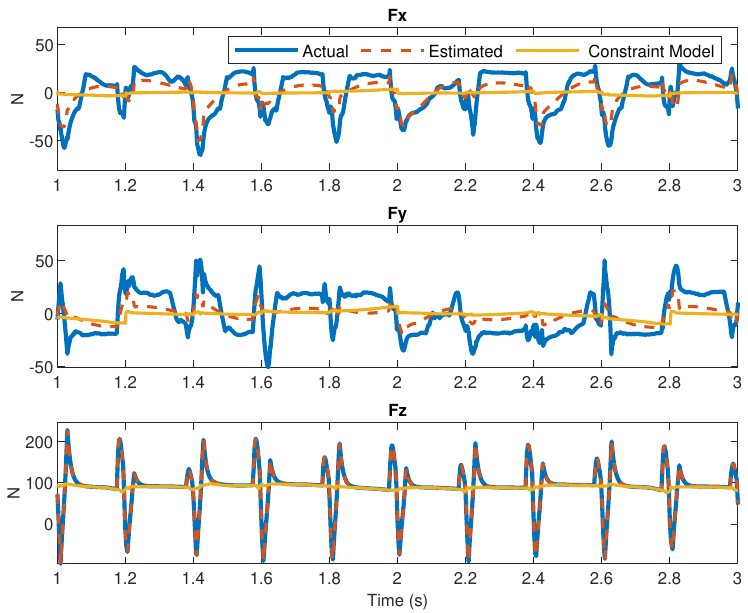}
\caption{Estimated Ground reaction forces from the momentum estimator compared with the constrained model. The momentum observer is able to track the estimated ground reaction forces for the reduced order model. The constrained order model cannot track the impulse forces experienced by the body. The gains used for the estimator are $\bm K_o = 1000*\mathbb{I}_6$}
\label{fig: Estimated forces}
\end{figure} 
Fig.~\ref{fig : body_states} show the evolution of the trajectory of the body position and attitude. With a slippery ground (\(\mu\) = 0.25 ), the ERG is able to efficiently able to find foot locations such that foot end doesn't slip. The constraint satisfaction is seen in Fig. ~\ref{fig : Friction cone constraint satisfication}. The controller is also able to find lateral footstep placement to account for roll of the robot ( See  Fig.~\ref{eq: leg states} and Fig.~\ref{fig: leg angles} . From this we can see how the ERG works well along with an operational space foot placement controller like the Raibert Heuristic. 
The momentum observer is able to observe the sum total of the ground reaction forces shown in Fig.~\ref{fig: Estimated forces}. The ground reaction forces are able to closely track the estimated ground reaction forces from the Spring-damper model. The Fig.~\ref{fig: Estimated forces} also shows the estimated ground reaction force from the constrained model is not able to track the ground reaction forces for the HROM. This could be due to consideration of massless legs and also that the foot contact Jacobian is not full rank when two legs are in contact with the ground. The estimator tracks the normal forces well (while also capturing the stiff spring-like impulsive behaviour) whereas the constrained model cannot see this as we consider a fixed foot end. The lateral forces are less accurate due to the usage of spring model and Coulomb friction for the ground model. 
\section{Conclusion}
In this study, we propose an optimization-free control framework, by filtering the applied reference, and ground contact force estimation framework for multimodal legged aerial robots. Both the controller and the estimator were running on a Matlab based numerical simulator. The simulation shows the ability of an optimization-free controller to enable thruster-assisted walking, and a conjugate momentum observer that is able to estimate ground contact forces and torques on the generalized coordinates. The simulation uses a reduced order model with massless legs and a spring model for the ground forces which could explain some of the errors in estimation, which were less pronounced compared to what we see from the constrained ground model. We expect to see better results with a more fleshed-out model by considering torque-controlled joints and even compliance at joint levels. To overcome the effect of the stiff ground model, a hybrid switching model with mode switching based on foot contacts could be used. A hybrid model would overcome the discontinuities obtained from a Coulomb model for the lateral forces. A further improvement can be seen if we use a stiff ODE-solver to tackle the stiff ground model where stiffness values exceed 10000 N/m

To take this even further, we would like to implement this on a high-fidelity simulation with state observers such as an EKF running to provide the conjugate momentum observer for ground contact force prediction. The full validation of this control and estimation framework would come from a hardware implementation on the Husky robot. Further research is required to see how we can use the obtained ground contact wrench to see if it can be used for predictive modeling. 
\printbibliography

@article{garone_explicit_2016,
	title = {Explicit {Reference} {Governor} for {Constrained} {Nonlinear} {Systems}},
	volume = {61},
	issn = {1558-2523},
	url = {https://ieeexplore.ieee.org/document/7244340/?arnumber=7244340},
	doi = {10.1109/TAC.2015.2476195},
	number = {5},
	urldate = {2024-10-01},
	journal = {IEEE Transactions on Automatic Control},
	author = {Garone, Emanuele and Nicotra, Marco M.},
	month = may,
	year = {2016},
	note = {Conference Name: IEEE Transactions on Automatic Control},
	pages = {1379--1384},
}

@inproceedings{gilbert_nonlinear_1994,
	title = {Nonlinear control of discrete-time linear systems with state and control constraints: a reference governor with global convergence properties},
	volume = {1},
	shorttitle = {Nonlinear control of discrete-time linear systems with state and control constraints},
	url = {https://ieeexplore.ieee.org/abstract/document/411031},
	doi = {10.1109/CDC.1994.411031},
	urldate = {2024-10-01},
	booktitle = {Proceedings of 1994 33rd {IEEE} {Conference} on {Decision} and {Control}},
	author = {Gilbert, E.G. and Kolmanovsky, I. and Tan, Kok Tin},
	month = dec,
	year = {1994},
	pages = {144--149 vol.1},
}

@article{bemporad_reference_1998,
	title = {Reference governor for constrained nonlinear systems},
	volume = {43},
	issn = {1558-2523},
	url = {https://ieeexplore.ieee.org/abstract/document/661611},
	doi = {10.1109/9.661611},
	number = {3},
	urldate = {2024-10-01},
	journal = {IEEE Transactions on Automatic Control},
	author = {Bemporad, A.},
	month = mar,
	year = {1998},
	note = {Conference Name: IEEE Transactions on Automatic Control},
	pages = {415--419},
}

@article{haddadin_robot_2017,
	title = {Robot {Collisions}: {A} {Survey} on {Detection}, {Isolation}, and {Identification}},
	volume = {33},
	issn = {1941-0468},
	shorttitle = {Robot {Collisions}},
	url = {https://ieeexplore.ieee.org/abstract/document/8059840?casa_token=f2FlWX9QHHwAAAAA:dBl8ndq2nJGpK2ZASHZs5jL6bGFcGvDbnKLnYPnZMRYBNLToFiwnABC-NNhfjzmuG6ZWlNTx},
	doi = {10.1109/TRO.2017.2723903},
	number = {6},
	urldate = {2024-09-29},
	journal = {IEEE Transactions on Robotics},
	author = {Haddadin, Sami and De Luca, Alessandro and Albu-Schäffer, Alin},
	month = dec,
	year = {2017},
	note = {Conference Name: IEEE Transactions on Robotics},
	pages = {1292--1312},
}

@inproceedings{de_luca_sensorless_2005,
	title = {Sensorless {Robot} {Collision} {Detection} and {Hybrid} {Force}/{Motion} {Control}},
	url = {https://ieeexplore.ieee.org/document/1570247},
	doi = {10.1109/ROBOT.2005.1570247},
	urldate = {2024-09-28},
	booktitle = {Proceedings of the 2005 {IEEE} {International} {Conference} on {Robotics} and {Automation}},
	author = {de Luca, A. and Mattone, R.},
	month = apr,
	year = {2005},
	note = {ISSN: 1050-4729},
	pages = {999--1004},
}

@inproceedings{flacco_residual-based_2016,
	title = {Residual-based contacts estimation for humanoid robots},
	url = {https://ieeexplore.ieee.org/document/7803308},
	doi = {10.1109/HUMANOIDS.2016.7803308},
	urldate = {2024-09-27},
	booktitle = {2016 {IEEE}-{RAS} 16th {International} {Conference} on {Humanoid} {Robots} ({Humanoids})},
	author = {Flacco, Fabrizio and Paolillo, Antonio and Kheddar, Abderrahmane},
	month = nov,
	year = {2016},
	note = {ISSN: 2164-0580},
	pages = {409--415},
}

@article{liu_sensorless_2024,
	title = {Sensorless {Ground} {Reaction} {Force} {Observation} {With} {Disturbance} {Compensation} in {Heavy}-{Legged} {Robots}},
	issn = {1941-014X},
	url = {https://ieeexplore.ieee.org/abstract/document/10423296?casa_token=08J0RNwSBz4AAAAA:ILZ0iM8HxUE14TJHfYH5xr924kKJ9mrpGnIXdvhzElwxCAdrid3dqij3JSw-KO8rPGVTIJuL},
	doi = {10.1109/TMECH.2024.3354989},
	urldate = {2024-09-27},
	journal = {IEEE/ASME Transactions on Mechatronics},
	author = {Liu, Shaoxun and Pan, Zheng and Zhou, Shiyu and Niu, Zhihua and Wang, Rongrong},
	year = {2024},
	note = {Conference Name: IEEE/ASME Transactions on Mechatronics},
	pages = {1--12},
}

@inproceedings{morlando_disturbance_2022,
	title = {Disturbance rejection for legged robots through a hybrid observer},
	url = {https://ieeexplore.ieee.org/abstract/document/9837169?casa_token=JNvn0qkC_8EAAAAA:5FKLrmvoZjz93PeCu--cGImCdfCACAF8q7d5-L2habWRzkP8CuabWbyDyveIBR0LxByRZspH},
	doi = {10.1109/MED54222.2022.9837169},
	urldate = {2024-09-27},
	booktitle = {2022 30th {Mediterranean} {Conference} on {Control} and {Automation} ({MED})},
	author = {Morlando, Viviana and Ruggiero, Fabio},
	month = jun,
	year = {2022},
	note = {ISSN: 2473-3504},
	pages = {743--748},
}

@inproceedings{lim_momentum_2021,
	title = {Momentum {Observer}-{Based} {Collision} {Detection} {Using} {LSTM} for {Model} {Uncertainty} {Learning}},
	url = {https://ieeexplore.ieee.org/document/9561667?denied=},
	doi = {10.1109/ICRA48506.2021.9561667},
	urldate = {2024-09-27},
	booktitle = {2021 {IEEE} {International} {Conference} on {Robotics} and {Automation} ({ICRA})},
	author = {Lim, Daegyu and Kim, Donghyeon and Park, Jaeheung},
	month = may,
	year = {2021},
	note = {ISSN: 2577-087X},
	pages = {4516--4522},
}

@inproceedings{vorndamme_collision_2017,
	title = {Collision detection, isolation and identification for humanoids},
	url = {https://ieeexplore.ieee.org/abstract/document/7989552},
	doi = {10.1109/ICRA.2017.7989552},
	urldate = {2024-09-17},
	booktitle = {2017 {IEEE} {International} {Conference} on {Robotics} and {Automation} ({ICRA})},
	author = {Vorndamme, Jonathan and Schappler, Moritz and Haddadin, Sami},
	month = may,
	year = {2017},
	pages = {4754--4761},
}

@misc{salagame_quadrupedal_2023,
	title = {Quadrupedal {Locomotion} {Control} {On} {Inclined} {Surfaces} {Using} {Collocation} {Method}},
	url = {http://arxiv.org/abs/2312.08621},
	urldate = {2024-04-05},
	publisher = {arXiv},
	author = {Salagame, Adarsh and Gianello, Maria and Wang, Chenghao and Venkatesh, Kaushik and Pitroda, Shreyansh and Rajput, Rohit and Sihite, Eric and Leeser, Miriam and Ramezani, Alireza},
	month = dec,
	year = {2023},
	note = {arXiv:2312.08621 [cs, eess]},
}

@inproceedings{dangol_hzd-based_2021,
	title = {A {HZD}-based {Framework} for the {Real}-time, {Optimization}-free {Enforcement} of {Gait} {Feasibility} {Constraints}},
	url = {https://ieeexplore.ieee.org/abstract/document/9555786?casa_token=PxYKdZY4ouoAAAAA:kEyJrAS0CR4AkGEGKqN8sBdQsRlUORvQFQkkw72eDWr8HU8bRzBslai8uvFHBCG91VQq6ayeV3Q},
	doi = {10.1109/HUMANOIDS47582.2021.9555786},
	urldate = {2023-12-09},
	booktitle = {2020 {IEEE}-{RAS} 20th {International} {Conference} on {Humanoid} {Robots} ({Humanoids})},
	author = {Dangol, Pravin and Lessieur, Andrew and Sihite, Eric and Ramezani, Alireza},
	month = jul,
	year = {2021},
	note = {ISSN: 2164-0580},
	pages = {156--162},
}

@inproceedings{sihite_efficient_2022,
	title = {Efficient {Path} {Planning} and {Tracking} for {Multi}-{Modal} {Legged}-{Aerial} {Locomotion} {Using} {Integrated} {Probabilistic} {Road} {Maps} ({PRM}) and {Reference} {Governors} ({RG})},
	url = {https://ieeexplore.ieee.org/abstract/document/9992754?casa_token=E0v5fTczBNsAAAAA:23fuUCarpuZwI_Fn00Mg5JZjwxGq8BB-i5n2y-rXXLdp14E3esNoXewTLtycARuiLaW-lmxIbWs},
	doi = {10.1109/CDC51059.2022.9992754},
	urldate = {2023-12-09},
	booktitle = {2022 {IEEE} 61st {Conference} on {Decision} and {Control} ({CDC})},
	author = {Sihite, Eric and Mottis, Benjamin and Ghanem, Paul and Ramezani, Alireza and Gharib, Morteza},
	month = dec,
	year = {2022},
	note = {ISSN: 2576-2370},
	pages = {764--770},
}

@article{sihite_multi-modal_2023,
	title = {Multi-{Modal} {Mobility} {Morphobot} ({M4}) with appendage repurposing for locomotion plasticity enhancement},
	volume = {14},
	copyright = {2023 The Author(s)},
	issn = {2041-1723},
	url = {https://www.nature.com/articles/s41467-023-39018-y},
	doi = {10.1038/s41467-023-39018-y},
	language = {en},
	number = {1},
	urldate = {2023-11-22},
	journal = {Nature Communications},
	author = {Sihite, Eric and Kalantari, Arash and Nemovi, Reza and Ramezani, Alireza and Gharib, Morteza},
	month = jun,
	year = {2023},
	note = {Number: 1
Publisher: Nature Publishing Group},
	pages = {3323},
}

@article{dangol_control_2021,
	title = {Control of {Thruster}-{Assisted}, {Bipedal} {Legged} {Locomotion} of the {Harpy} {Robot}},
	volume = {8},
	issn = {2296-9144},
	url = {https://www.frontiersin.org/articles/10.3389/frobt.2021.770514},
	urldate = {2023-11-18},
	journal = {Frontiers in Robotics and AI},
	author = {Dangol, Pravin and Sihite, Eric and Ramezani, Alireza},
	year = {2021},
}

@inproceedings{liang_rough-terrain_2021,
	title = {Rough-{Terrain} {Locomotion} and {Unilateral} {Contact} {Force} {Regulations} {With} a {Multi}-{Modal} {Legged} {Robot}},
	doi = {10.23919/ACC50511.2021.9483189},
	booktitle = {2021 {American} {Control} {Conference} ({ACC})},
	author = {Liang, Kaier and Sihite, Eric and Dangol, Pravin and Lessieur, Andrew and Ramezani, Alireza},
	month = may,
	year = {2021},
	note = {ISSN: 2378-5861},
	pages = {1762--1769},
}

@inproceedings{sihite_optimization-free_2021,
	title = {Optimization-free {Ground} {Contact} {Force} {Constraint} {Satisfaction} in {Quadrupedal} {Locomotion}},
	doi = {10.1109/CDC45484.2021.9683155},
	booktitle = {2021 60th {IEEE} {Conference} on {Decision} and {Control} ({CDC})},
	author = {Sihite, Eric and Dangol, Pravin and Ramezani, Alireza},
	month = dec,
	year = {2021},
	note = {ISSN: 2576-2370},
	pages = {713--719},
}

@inproceedings{sihite_dynamic_2024,
	title = {Dynamic modeling of wing-assisted inclined running with a morphing multi-modal robot},
	url = {https://ieeexplore.ieee.org/document/10610678},
	doi = {10.1109/ICRA57147.2024.10610678},
	urldate = {2024-08-26},
	booktitle = {2024 {IEEE} {International} {Conference} on {Robotics} and {Automation} ({ICRA})},
	author = {Sihite, Eric and Ramezani, Alireza and Gharib, Morteza},
	month = may,
	year = {2024},
	pages = {2339--2345},
}

@inproceedings{krishnamurthy_narrow-path_2024,
	title = {Narrow-{Path}, {Dynamic} {Walking} {Using} {Integrated} {Posture} {Manipulation} and {Thrust} {Vectoring}},
	url = {https://ieeexplore.ieee.org/document/10637015},
	doi = {10.1109/AIM55361.2024.10637015},
	urldate = {2024-08-26},
	booktitle = {2024 {IEEE} {International} {Conference} on {Advanced} {Intelligent} {Mechatronics} ({AIM})},
	author = {Krishnamurthy, Kaushik Venkatesh and Wang, Chenghao and Pitroda, Shreyansh and Salagame, Adarsh and Sihite, Eric and Nemovi, Reza and Ramezani, Alireza and Gharib, Morteza},
	month = jul,
	year = {2024},
	note = {ISSN: 2159-6255},
	pages = {898--903},
}

@inproceedings{ramezani_generative_2021,
	title = {Generative {Design} of {NU}’s {Husky} {Carbon}, {A} {Morpho}-{Functional}, {Legged} {Robot}},
	url = {https://ieeexplore.ieee.org/abstract/document/9561196},
	doi = {10.1109/ICRA48506.2021.9561196},
	urldate = {2023-11-22},
	booktitle = {2021 {IEEE} {International} {Conference} on {Robotics} and {Automation} ({ICRA})},
	author = {Ramezani, Alireza and Dangol, Pravin and Sihite, Eric and Lessieur, Andrew and Kelly, Peter},
	month = may,
	year = {2021},
	note = {ISSN: 2577-087X},
	pages = {4040--4046},
}

@article{dangol_towards_2020,
	series = {21st {IFAC} {World} {Congress}},
	title = {Towards thruster-assisted bipedal locomotion for enhanced efficiency and robustness},
	volume = {53},
	issn = {2405-8963},
	url = {https://www.sciencedirect.com/science/article/pii/S2405896320334844},
	doi = {10.1016/j.ifacol.2020.12.2721},
	language = {en},
	number = {2},
	urldate = {2023-05-17},
	journal = {IFAC-PapersOnLine},
	author = {Dangol, Pravin and Ramezani, Alireza},
	month = jan,
	year = {2020},
	pages = {10019--10024},
}
\end{document}